\documentclass[runningheads]{llncs}
\usepackage{graphicx}

\usepackage{float}
\usepackage[square,numbers]{natbib}
\usepackage[utf8]{inputenc}
\usepackage{xcolor}
\usepackage{longtable}
 
\usepackage{amsmath}
\DeclareMathOperator*{\argmaxA}{arg\,max}

\begin{document}
\title{Survey of XAI in digital pathology\thanks{This work was supported by the Swedish e-Science Research Center.}}
%
%
\author{Milda Pocevičiūtė\inst{1,2} \and
Gabriel Eilertsen \inst{1,2} \and
Claes Lundström\inst{1,2,3}}
\authorrunning{M. Pocevičiūtė et al.}
%
\institute{Department of Science and Technology, Linköping University, Sweden\\
\email{\{milda.poceviciute, gabriel.eilertsen, claes.lundstrom\}@liu.se} \and
Center for Medical Image Science and Visualization, Linköping University, Sweden \and
Sectra AB, Linköping, Sweden}
\maketitle              
\begin{abstract}
Artificial intelligence (AI) has shown great promise for diagnostic imaging assessments. However, the application of AI to support medical diagnostics in clinical routine comes with many challenges. The algorithms should have high prediction accuracy but also be transparent, understandable and reliable. Thus, explainable artificial intelligence (XAI) is highly relevant for this domain. We present a survey on XAI within digital pathology, a medical imaging sub-discipline with particular characteristics and needs. The review includes several contributions. Firstly, we give a thorough overview of current XAI techniques of potential relevance for deep learning methods in pathology imaging, and categorise them from three different aspects. In doing so, we incorporate uncertainty estimation methods as an integral part of the XAI landscape. We also connect the technical methods to the specific prerequisites in digital pathology and present findings to guide future research efforts. The survey is intended for both technical researchers and medical professionals, one of the objectives being to establish a common ground for cross-disciplinary discussions.

\keywords{XAI \and digital pathology \and AI \and medical imaging.}
\end{abstract}

\section{Introduction and Motivation}
\subsection{Background}
Artificial intelligence (AI) applications are showing great promise for assisting diagnostic tasks in medical imaging. Nevertheless, it is difficult to translate the technology from academic experiments to clinical use. A central challenge for AI in medicine is that mistakes can have serious consequences. This means that human experts must be able to gauge the trustworthiness of machine predictions, and put it into the context of other diagnostic information. This is the purpose of explainable artificial intelligence (XAI) techniques. XAI research embraces the insight that AI solutions should not only have high accuracy performance, but also be transparent, understandable and reliable from the end user's point of view. 
\par
This survey investigates XAI in the domain of digital pathology. The adoption of digital microscopy, whole-slide imaging (WSI), at clinical pathology departments is progressing at a fast pace in many parts of the world. A key motivation for this investment is the potential to use AI assistance for image analysis tasks. XAI has been described as an essential component to make AI successful in imaging diagnostics \cite{Litjens2017Survey}, and we argue this is particularly pertinent for digital pathology. For example, assume that a pathologist is faced with an AI result marking a WSI region as “benign tumour”, whereas the pathologist deemed it as probably malignant. It is easy to see how the pathologist would need further information on the rationale of the machine prediction in order to accept or reject the result that was in conflict with his/her own initial assessment.
\par
There are several motivations to specifically target XAI for digital pathology, as we do in this survey. XAI has so far been dominated by the explainability tailored for AI developers, whereas the needs of pathologists and other medical professionals are distinctly different, as will be described below. Pathology is also quite different from other medical imaging. Gigapixel images are the norm, which are visually scrutinised on many scales. The characteristics of the large histology ``landscapes" are very different both from photos such as in ImageNet and from radiological images. We believe that describing the XAI prerequisites in pathology will be valuable for informing much needed future research efforts in this domain.
\par
This survey is a niched drill-down complementing previous more general reviews. There are broad XAI overviews \cite{Adadi2018PeekingIT,Dosilovic2018,goebel2018,Hohman2019PathReview,mueller2019explanation}, and more specialised reviews such as for Convolutional Neural Network (CNN) methods \cite{ZhangQuan2018}. A few efforts discuss the potential of XAI for medicine in general \cite{Holzinger2019,Tjoa2019ASO}. 
\par
There are several specific contributions in our survey. We have elicited a classification of XAI techniques from three separate aspects: \emph{explanation target}, \emph{result representation}, and \emph{technical approach}. We have identified previous XAI efforts with relevance for digital pathology, most of them not applied to this domain yet, and categorise them into the defined classes. Estimation and visualisation of uncertainty are sometimes treated as a topic separate from explainability. We echo previous researchers arguing against such a separation \cite{alvarez-melis-jaakkola-2017-causal, Holzinger2019, Tagasovska2018} and incorporate uncertainty methods as an inherent part of the XAI overview in this review. Finally, based on an analysis of the survey outcome, we provide some key findings that could guide future research efforts to solve XAI challenges in digital pathology. We believe that this paper is suitable for both technical researchers and medical domain professionals. For example, the categorisation is made with both target groups in mind, where \emph{result representation} and \emph{explanation target} are of interest to the medical experts, whereas the \emph{technical approach} is separated into an isolated group. Thus, we believe that the survey can assist in understanding across the disciplines by providing a joint structure as a base for discussions. 
\par
Our survey places the focus on image recognition tasks as most AI algorithms in digital pathology work with image data. Therefore, all methods described in this survey are applicable for CNN models as, currently, this is the state-of-the-art in digital pathology. We use the terms AI tools, AI solutions and AI algorithms interchangeably.

\subsection{AI in pathology}

The workload for pathologists is predicted to increase continuously due to the ageing population, shortage of physicians, increased cancer screening programmes and increased complexity of diagnostic tests \cite{Serag2019PatReview}. One way of addressing this problem is to introduce digital pathology; that is, new workflows and tools based on the digitisation of microscopy images \cite{Thorstenson2014}. The possibility to add assistive AI tools is a major component of the foreseen improvements. As a foundation for our discussion on XAI, we will in this section first provide a brief overview of some important types of AI use cases for the clinical routine setting of digital pathology. For more exhaustive overviews of applied AI research in this area, we refer to previous review efforts \cite{Bera2019,Hohman2019PathReview,Serag2019PatReview}, and for an introduction to the diagnostic work of a pathologist, we refer to \cite{Pohn2019}.
\par
A common diagnostic task in pathology is to detect the existence of cancer. Thus, AI development efforts are often directed towards assisting the tumour detection process. One improvement aspect is to make the search more efficient. Since the lesions may be just a handful of cells in a sea of normal tissue, it can be very time-consuming to locate them. For some scenarios the search can stop when a first lesion is found, meaning that normal/benign cases are the most time-consuming as the search then covers the entire sample. Metastasis detection in breast cancer lymph nodes is a common AI research application \cite{Bejnordi2017,LiuGadepalli2017}. The other task aspect is to determine whether a finding actually is malignant or not, and often this includes performing a subtype classification of the cancer in question. Illustrative Cresearch efforts in this subarea include a tool for detection and subtype classification of gliomas and non-small-cell lung carcinomas \cite{HouSamaras2016}, classification of gastric carcinoma \cite{SHARMA20172}, and malignancy prediction in breast cancer cytology \cite{GarudKarri2017}.
\par 
Detection tools could also help to reorganise the worklist of a pathologist so that the cases with a high risk of malignant tumours would be prioritised. Apart from tumours, potential detection tasks for AI include needle-in-a-haystack searches for tuberculosis or helicobacter pylori bacteria \cite{molin2016scale}.
\par
In the diagnostic work-up of oncology cases, the pathologist typically provides further granularity in the analysis in the form of grading and staging assessments. These assessments often suffer from poor inter-observer reproducibility as well as high time consumption, making AI assistance attractive. In breast cancer, quantification of cell proliferation is part of the grading. Detecting and counting mitotic cells is a common target for AI methods \cite{Balkenhol2019,ChenDou2016,MitkoYujing2018}. AI solutions are commonly employed also for other cell quantification in breast cancer diagnostics, regarding positive nuclei in sections stained through immunohistochemistry (IHC) \cite{Alzubaidi2017,Hofener2018,Jung2019,narayanan2018deepsdcs,XueRay2016}. Quantified IHC analysis is relevant to predict response for many targeted treatments, with active research efforts in AI method development. Important examples include detection of positive cell membranes in the PD-L1 \cite{Koelzer2018} and HER2 \cite{Saha2018} stains.
\par
The Gleason score is used to stage prostate cancer by assessing the extent of different architectural patterns of a tumour. This analysis has been in focus for applied AI research \cite{Arvaniti2018,Nagpal2019} and recent larger studies show results on par with human experts \cite{Bulten2020, Strom2020}. 
\par
Another cell identification task is to count lymphocytes, which for example is important for predicting treatment response for immunotherapy on cancer patients. Deep learning methods have shown the potential to provide support for this diagnostic task as well \cite{chen2016automatic,Garcia-Rios2017,SWIDERSKACHADAJ2019101547}.
\par
A pathologist's assessment can be underpinned by other more generic quantification where AI could contribute to higher efficiency. This is relevant for area measures such as tumour/stroma ratio or tumour necrosis rate, and potentially for automation of distance measurements such as tumour thickness in skin cancer and the margin from the tumour to the resection border.  
\par
\par
Apart from the tasks described above, neural networks have the potential to be used for content-based image retrieval (CBIR) to find previous cases with similar histology patterns as the patient at hand. This can not only assist the daily work of a pathologist but also improve the education of new physicians \cite{KomuraIshikawa2017}. Deep learning has been employed for this purpose, both in research \cite{CaiReif2019} and in a commercial application \cite{Huron}. 
\par
Overall, deep learning is a flexible approach that can be used to assist pathologists in many different aspects of their work. However, the path from promising AI models to actual clinical usage is very challenging. We argue that a key part of meeting that challenge is to develop effective and tailored XAI methods. Next, we will drill down into the specific needs of XAI in this domain.

\subsection{Needs of XAI in digital pathology} \label{Challenges}
\begin{figure}[t]
\centering
\includegraphics[width=\textwidth]{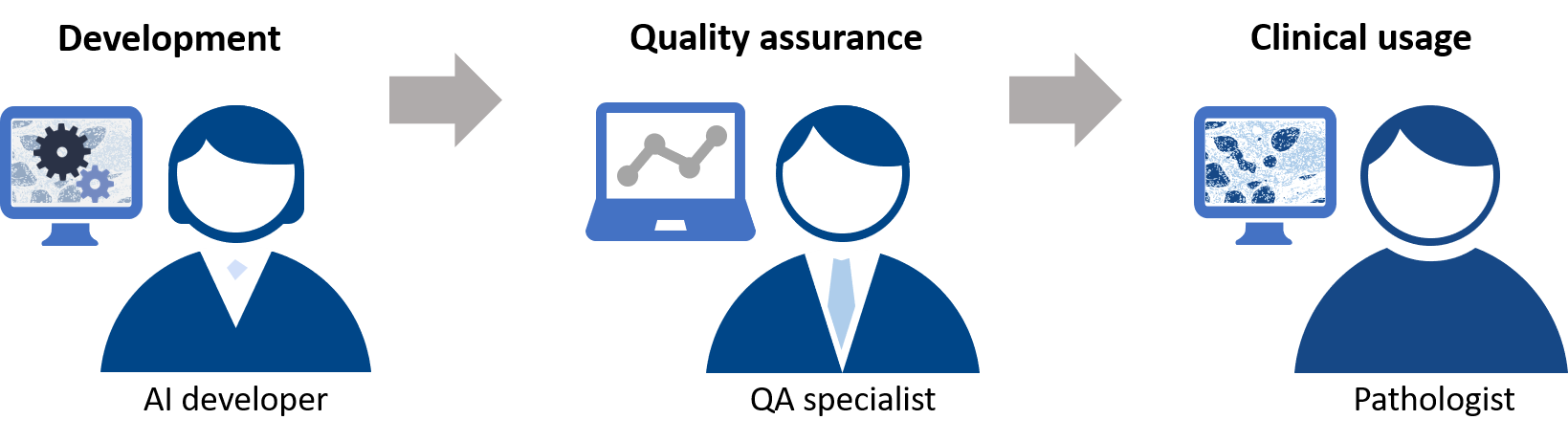}
\caption{Overview of the three main scenarios in digital pathology where XAI methods are relevant.}
\label{fig:aichain}
\end{figure}

XAI serves different purposes depending on the role of the person receiving the explanation, and depending on the reason for interacting with the AI technology. In digital pathology for clinical use we see three main scenarios, having quite different characteristics (Figure~\ref{fig:aichain}). The arguably most common target for XAI is to assist the AI developer working to create or improve the model, which of course is relevant also in this domain. The second scenario is when the clinical end user, typically the pathologist, employs an AI solution in the diagnostic routine. The third XAI target area, perhaps less considered than the others, is healthcare professionals doing quality assurance (QA) of AI solutions. This role may be taken by pathologists or other medical staff, but we may also see data scientists entering the diagnostic departments to carry out such assignments. QA can correspond to initially assessing how well an algorithm performs at the lab in question, for calibrating or configuring the solution to fit local characteristics, to evaluate if there is a drift in performance over time, and more.

\par
The AI developer perspective on XAI needs is fairly generic, at the conceptual level it is the same in digital pathology as in other application areas. We will give an outline here, while further details can be found in the survey by Hohman et al. \cite{Hohman2019PathReview}. Explainability is a key component to support the developer in improving the performance of the trained model. For example, studying erroneous predictions is effective for evaluating the next development steps to be taken. Prediction accuracy aside, the developer also benefits from XAI analysing the generalisability of the results; is the training data, including augmentations made, representative for wider use, and is there any bias to consider? As collecting and preparing data could be very laborious, often including human experts spending many hours annotating images, the collection may not have sufficient coverage for the intended application. For instance, if there are many classes to separate, some of them may have too few examples in the data collected. 
\par
For the routine diagnostic use of AI solutions, there are many situations where explainability would be beneficial. In fact, we argue that effective XAI is essential for broad successful deployment of computational pathology in the clinical environment. 
\par
For a physician using AI assistance, a main task is to spot and correct any errors made by the algorithm. XAI would then provide assistance to critically assess the result. The typical ML model always predicts one of its predetermined outcomes, even if the evidence is lacking for any of the alternatives. An important aspect is therefore to convey the uncertainty of the prediction made. This is particularly useful when there is not a black-or-white assessment to be made, but a conclusion from a complex set of contributing factors. 
\par
The physician would likely also benefit from a deeper understanding of the source of the AI tool's limitations in the context it is used. For models trained by supervised learning, the representativeness of the training data in relation to the local data characteristics is a key factor, i.e. its ability to generalise. Image differences due to e.g. staining variations is a well-known challenge in computational pathology \cite{Stacke2019, TELLEZ2019}, creating a domain gap between the training data and the data used for inference. There may also be discrepancies in the definition of the diagnostic task trained for and the local diagnostic protocol. Such problems makes it important to highlight when the diagnosis provided by an AI application cannot be trusted due to a lack in its ability to generalise to the current situation.
\par
Achieving the above transparency is useful for making correct conclusions for individual cases, but also to allow the medical professionals to gain trust in the solution in general. Such trust is necessary in order to arrive at an effective division of labour between man and machine. Powerful XAI can, however, induce too high levels of trust, counteracting the objective of critical assessment \cite{Kaur2020}. Therefore, XAI methods should be carefully designed to provoke sound reflection, rather than just creating blind trust.
\par
An additional benefit of explainable AI predictions is for teaching. Whereas it today is very difficult to verbalise the complex assessments a senior pathologist does \cite{Holzinger2019}, XAI visualisations may be able to better convey such knowledge. Furthermore, there is a direct connection to medical research, as XAI may help uncover previously unknown disease characteristics.
\par
In digital pathology, AI researchers face the challenge of dealing with very large data sets, typically gigapixel images. Diagnostic pathology assessments almost always include considering small-scale and large-scale features in tandem, and computational pathology needs to do the same. This is a particular challenge for XAI as well. An AI prediction will likely need to be explained both at local cellular level and at a higher tissue structure level.
\par
The QA scenario shares many of the XAI needs of the diagnostic assessments described above. Identifying errors, assessing uncertainty, and understanding limitations are in focus here as well. The difference is that the focus shifts from individual cases to analyses across larger case sets representing the entire operations at the lab. This poses additional requirements for XAI solutions, to support systematic, in-depth investigations akin to the diligent validation procedures for other lab equipment.
\par

The aspects discussed above clearly shows that there is a great diversity of situations where pathology diagnostics require transparency and interpretability. In summary, there is strong rationale for XAI advances tailored for digital pathology.

\section{Glossary} \label{Pathology}
\textbf{XAI}: explainable artificial intelligence, a field of study on how to increase transparency and intepretabily of artificial intelligence algorithms such that the results could be understood by a human expert.
\par
\noindent \textbf{Standard neural network (NN)}: the most common type of neural network used in research as well as in practical applications. They are based on frequentist probability theory which states that each parameter of a model has a true fixed value. It is hard and often impossible to find the exact true values, hence the backpropogation algorithm provides a means of approximating them \cite{Goodfellow2016}. In the context of the uncertainty estimation, the parameters of a NN are not random variables, hence, probabilities cannot be associated with them and other frequentist techniques have to be used \cite{YoungStatisticalInference}.
\par
\noindent \textbf{Bayesian neural network (BNN)}: a type of neural network that is based on Bayesian theory and requires more complex procedures of training and inference. According to the Bayesian approach, a probability is a degree of the belief that a certain event will occur and can be modelled by Bayes' theorem. Bayes' theorem states that the conditional probability of an event depends on the data as well as prior information/belief. The model parameters are viewed as random variables, and Bayesian inference is used to estimate the probability distribution over them. This distribution reflects the strength of the belief regarding what parameter values that are possible, and it is used during forward pass through a BNN to sample some likely parameters and produce a range of possible outcomes \cite{YoungStatisticalInference}.

\section{State-of-the-art} \label{TypesNN}
In this section, we provide an overview of the current methods that aim to help in developing transparent and interpretable AI solutions. We focus on methods that are specifically developed for, or can be easily applied to, visual detection tasks. There are many ways in which the methods can be grouped. In this work, we provide three alternative taxonomies, namely \emph{Explanation target}, \emph{Result representation} and \emph{Technical approach}. Each taxonomy is described in detail in the corresponding subsection below, followed by examples of representative XAI methods that can be assigned to it. Table \ref{tab:methods_cats} in Appendix \ref{sec:appendix} contains all reviewed XAI methods, classified according to the three alternative ways of categorisation. 
\par
The explanation target gives a general understanding of what can be explained in visual detection tasks and for which group of professionals -- AI developers, QA specialists or pathologists -- this explanation is most relevant. Result representation illustrates how the explainability may be presented in an AI solution while the Technical approach provides an insight into what techniques and mathematical theories are used in order to achieve the explainability. Many of the existing XAI techniques focus on explaining classification tasks, or at least illustrate their work with examples of classification algorithms. Some methods encompass other computer vision tasks, such as detection, localisation and segmentation. Furthermore, it is important to note that some methods can be or have already been adapted for different tasks which could make them fall under several different categories. We decided to base our categorisation on how the method is described in the original paper. 
\par
\begin{figure}[t]
\centering 

\includegraphics[width=\textwidth]{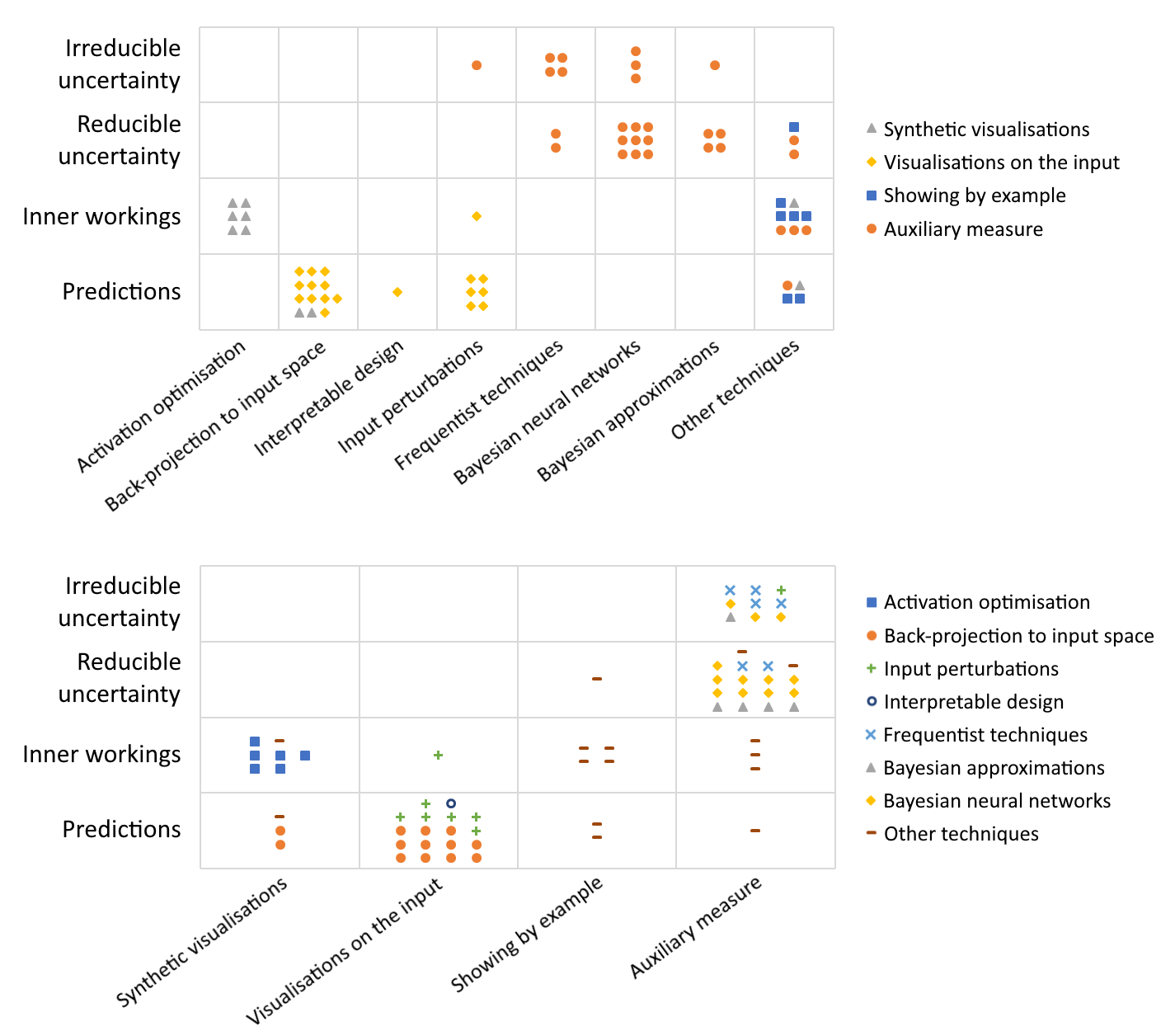}
\caption{Summary of the reviewed papers based on the three ways of categorisation. Explanation targets are given on the y-axis. The top plot has Technical approach as x-axis with markers indicating different Result representations. Conversely, the bottom plot shows Result representations on the x-axis with the markers indicating the Technical approach. Papers containing two method components representing different categories have been registered twice.}
\label{fig:SummaryMeths}
\end{figure}

\begin{figure}[t]
\centering
\includegraphics[width=0.7\textwidth]{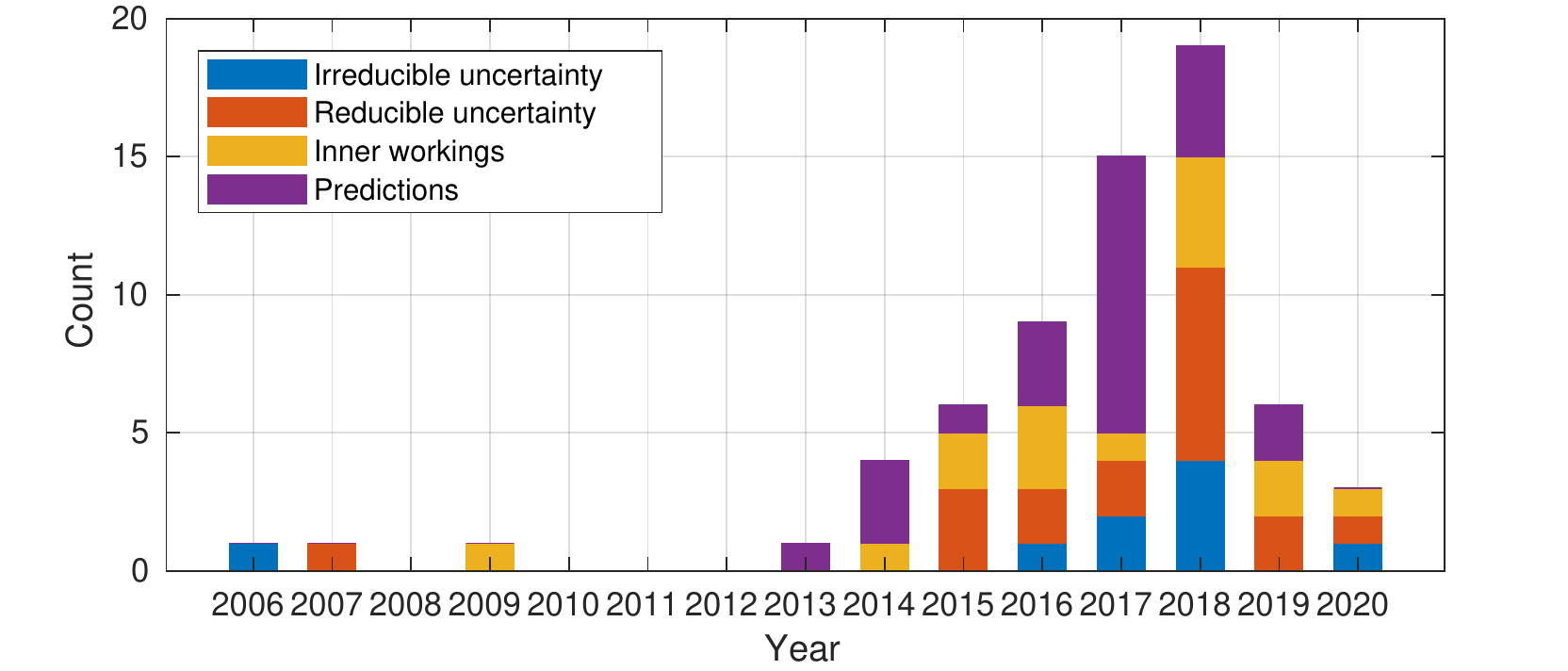}
\caption{Time histogram of the methods in Figure~\ref{fig:SummaryMeths}. Grouping is made according to the different explanation targets.}
\label{fig:time_hist}
\end{figure}

 
Figure \ref{fig:SummaryMeths} summarises the reviewed papers based on our three dimensions of categorisation, and Figure \ref{fig:time_hist} shows the development over time. In Figure \ref{fig:SummaryMeths}, the matrix plot in the top gives an overview of what technical approaches are most commonly used for which explanation targets. In contrast, the plot in the bottom gives an overview of what result presentation types are most commonly used for which explanation targets. We can see that irreducible uncertainty so far has been only presented as an auxiliary measure even though there are quite a few different technical approaches for determining it. In contrast, the explanation of inner workings can be presented in many different ways, but the activation optimisation approach is the most commonly used to achieve the results. Figure \ref{fig:SummaryMeths} not only summarises previous work in XAI but also highlights which combinations of the categories that have not yet been explored. 
\par 
It is important to note that this section is aimed at providing a general understanding of existing methods, hence the text does not focus on the pathology specific aspects. However, the result representation part is mainly illustrated by XAI methods applied to histopathological data. Furthermore, a discussion on how the different methods can be used for fulfilling the need for reliable and transparent AI tools of digital pathology is provided in Section \ref{Discussion}. 

\subsection{Explanation target}
\begin{figure}[t]
\centering
\includegraphics[scale = 0.3]{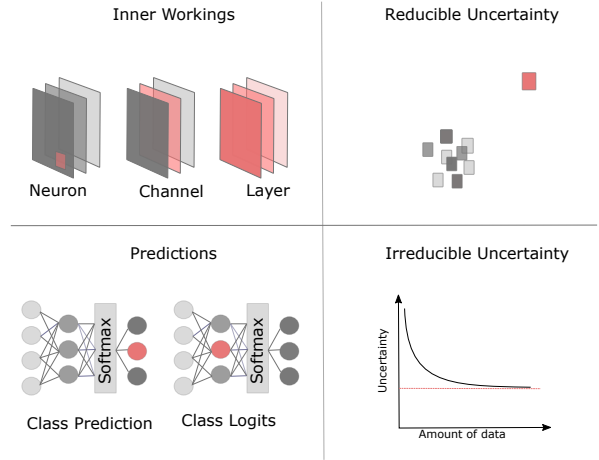}
\caption{Illustration of the different explanation targets. In order to understand inner workings, we can explore neurons, channels, layers and the relations between them. Explaining the predictions only focus on a specific class output and what logic the model is using to arrive at them. Understanding reducible uncertainty enables us to detect outliers in the data and warn when an NN is making an uninformed prediction, while the irreducible uncertainty shows how well our model approximates the variability in the phenomenon of interest.}
\label{fig:PredInner}
\end{figure}

Explainability could have several objectives in the context of AI assisting visual analysis. Figure \ref{fig:PredInner} illustrates the four targets that an XAI method may help to understand better. In this section we describe in more detail each of the explainability types and illustrate with some examples from the reviewed papers.

\subsubsection{Explaining predictions} \label{ExplPred}
Explaining predictions refers to methods that are developed to understand why a certain image led to a corresponding NN output. The expectation is often that the reasoning of an NN would match the logic that human experts use when solving the problem. For example, if an NN is detecting the presence of a tumour, we would like to see that the NN is basing its prediction on the areas of the image that contain tumour cells. This would reassure pathologist that the prediction is trustworthy. Some well-known examples in this category include saliency maps \cite{Simonyan2013}, Excitation Backprop \cite{Zhang2018ExcitationBackprop} and explanations generated using meaningful perturbations \cite{Fong2017}. Such explanations could be useful for the AI developers to debug their created algorithms, could aid QA specialists in guaranteeing an appropriate behaviour of an NN, and could foster trust in the community of the end-users. 

\subsubsection{Explaining inner workings} \label{ExplInner}
Methods in the category of inner workings explanation aim to increase the understanding of how an algorithm works. There are a few ways that researchers have attempted to achieve this. For example, activation maximisation techniques show the patterns to which neurons react most \cite{Erhan2009, Nguyen2016, yosinski2015understanding}. Other methods analyse image representations encoded by the model in order to understand what information is retained and what is lost. This can be done by recreating the original images from the representations saved at different layers of a CNN \cite{Dosovitskiy2016,Mahendran2015}. Finally, some techniques examine what interpretable concepts, such as textures and patterns, that are detected for a specific class of interest. This enables explanation of behavioural properties of deep neural networks \cite{Kim2018,Leino2019}. This explanation type is most relevant to the AI developers as it can give new insights into what an NN is doing and how its performance could be boosted.

\subsubsection{Understanding reducible uncertainty}  

Reducible uncertainty, also known as epistemic uncertainty, refers to when the training of the model has been imperfect, insufficiently covering the phenomenon of the interest. Having high such uncertainty means that the parameters of an NN are not tuned properly to make an informed prediction for some inputs \cite{DerKiureghian2008}, i.e., data points being outliers in relation to the training data. This results in the model being incapable of providing informed predictions about outliers as the prior knowledge gathered during training is insufficient. If we would expand the training data and ensure it contains all outliers, this uncertainty could be reduced to zero. Even without more data, an improved training scheme could also reduce the epistemic uncertainty. Explanations targeting this uncertainty type enable us to understand the limitations of our model via training data and how we can improve it to increase prediction accuracy.
\par
There are two main ways of estimating the epistemic uncertainty. It can be modelled on the NN's weights directly \cite{Blundell2015,Hernandez-Lobato2015,Ritter2018}. Other methods use the exploration of the data sets used for training and validating an NN to find out which points that are outliers and tune the uncertainty estimates to match this information \cite{Hafner2019,Papadopoulos2007,Papernot2018}.

\subsubsection{Understanding irreducible uncertainty}
The irreducible uncertainty arises due to the intrinsic randomness of a phenomenon of interest and the fact that our model is an approximation of it. It is is also known as an aleatoric uncertainty  \cite{DerKiureghian2008}. This uncertainty cannot be reduced by increasing the size of the training data set. For an intuitive understanding, consider that even if we train a model to detect a tumour in WSIs on all of the relevant images existing in the world, our model is still an approximation of the complex phenomenon (tumour in human tissue). Hence, the upcoming new images from new patients may inherit some variability that our model is unable to classify correctly due to the missing variables/parameters that are necessary for capturing it. 
\par
So what does such uncertainty explain to us? It reminds the users that they should never trust an AI prediction completely as the nature of the world contains intrinsic variations that none of the models can perfectly capture. Furthermore, aleatoric uncertainty also gives an insight into whether developers chose an appropriate architecture of the model for a particular problem. If the developers observe low epistemic uncertainty but high aleatoric uncertainty, this indicates that the model is too simple to approximate the properties of the phenomenon of the interest well \cite{KhanImmer2019}. In an ideal situation, the chosen model would only contain low aleatoric uncertainty and the prediction variability would be as low as possible. Some examples of methods for estimating the aleatoric uncertainty include \cite{Ayhan2018, Bouchacourt2016,Gal2016}. 

\subsection{Result representation}
The categorisation of result representations assesses what type of results the user should expect to receive from applying an explainability technique. This allows users and developers to quickly pick a subgroup of methods that would provide the desired explanation output. We have distinguished four main groups of how the results are presented: \emph{synthetic visualisations}, \emph{visualisations on an input image}, \emph{showing by example} and \emph{auxiliary measures}. If we are working on helping a pathologist with a tumour detection task or want to boost the pathologist's confidence in an AI solution, we may be most interested in the techniques that provide visualisations on the original input images or show how an NN works by an example. However, if we are debugging or validating the overall classification strategy of an NN, the methods that generate synthetic visualisations or auxiliary measures may be preferred.

\subsubsection{Synthetic visualisations}

\begin{figure}[t]
\centering
\includegraphics[width=\textwidth]{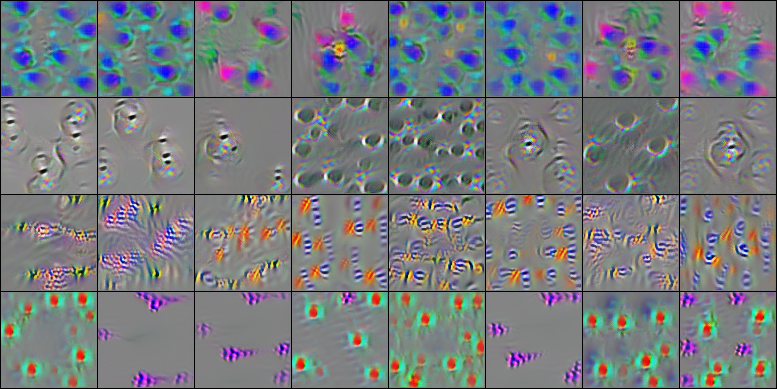}
\caption{Activation maximisation of GoogLeNet filters (per channel) trained on histopaphology data. Each row corresponds to filters from the same model, but trained on different formulations of the same original dataset, from top to bottom: unmodified original data, colour and intensity augmentations, stain normalisation, and Cycle-GAN transformation for domain-adaptation between different medical centres. Reproduced with the authors' permission from \cite{Stacke2019}.}
\label{fig:ActivationMax}
\end{figure}

Synthetic visualisations is a broad group of methods that all generate a synthetic image as an outcome. The group can be further divided based on what sort of image is generated.
\par
The first subgroup generates surrealistic images from scratch that can be interpreted as illustrations of what some part of the NN reacts to. These methods are also known as activation maximisation: they attempt at finding, through optimisation, the input images that maximally activate a chosen neuron, channel or layer. Some examples of the methods can be found in \cite{Carter2019, Erhan2009, Nguyen2016, Olah2018, yosinski2015understanding}. The visualisations give an insight into what patterns the neurons are looking for. For example, this knowledge may aid the analysis of a domain shift problem \cite{Stacke2019}. Figure \ref{fig:ActivationMax} shows an example of patterns that maximally activate some filters of a Mini-GoogLeNet that is trained to classify tumours in WSI patches \cite{Stacke2019}. The different rows use different strategies for formulating the training data but are all based on the same original dataset. The results show how very different representations are learned by the convolutional layers depending on how the data is pre-processed, e.g. how colour augmentation makes the representation less sensitive to absolute colour values in the input (second row).
\par
Another subgroup of methods focuses on using the feature representations in an NN to generate images that resemble parts of the original input. The aim is not to retrieve the original image but to conclude which patterns could be most responsible for the NN’s prediction (or a neuron’s activation). A well-known method of this category is deconvolutional networks \cite{ZeilerFergus2014}. Other examples that aim to explain the target NN include PatternNet \cite{Kindermans2017} and Guided Backprop  \cite{Springenberg2014}. The main drawback these methods have is that they produce visualisations that are not class-specific. This means that the techniques give an insight into which patterns are important in general, but cannot be used to understand why an NN predicted the specific class \cite{Nie2018ATE}.
\par
Finally, some methods have focused on reconstructing the original images from the feature representations in an NN. They are also known as model inversion methods. This type of visualisation illustrates what information an NN is keeping about the image which, in turn, helps to understand better how an NN works. For example, comparing the reconstructed images to the original input, Dosovitskiy and Brox  \cite{Dosovitskiy2016} found that the colour seems to be a crucial feature for the prediction they studied. Similar visualisations can be found in \cite{Mahendran2015, Mahendran2016}.

\subsubsection{Visualisations on an input image} \label{Heatmaps}

The methods in this group produce three types of visualisations: heatmaps, important patches and receptive fields. A heatmap is a graphical representation of 2D scalar data where the values are encoded by colour. With such representation, this first type of visualisation shows how much each pixel contributes to the prediction. This information is visually conveyed by overlaying the colour gradient on the original input image.  Well-known techniques that produce heatmaps are Excitation Backprop \cite{Zhang2018ExcitationBackprop}, Grad-CAM \cite{Selvaraju2017} and Layer-wise relevance propagation \cite{Bach2015}. The second type of visualisation is produced by keeping the important patches (pixel regions) of the original input and cropping out the remaining. These techniques reveal which objects or regions in the input that are contributing to the prediction. However, they do not provide the knowledge of importance distribution over the pixels  \cite{Ribeiro2016, Zhou2014}. The third type of visualisation marks the receptive field, the areas on the original input image that indicate what regions that activate most a target unit \cite{ZhangWuZhu2018,Zhou2014}. This gives an insight into how neurons in NNs and filters in CNNs work. 

\begin{figure}[t]
\centering
\includegraphics[scale=0.4]{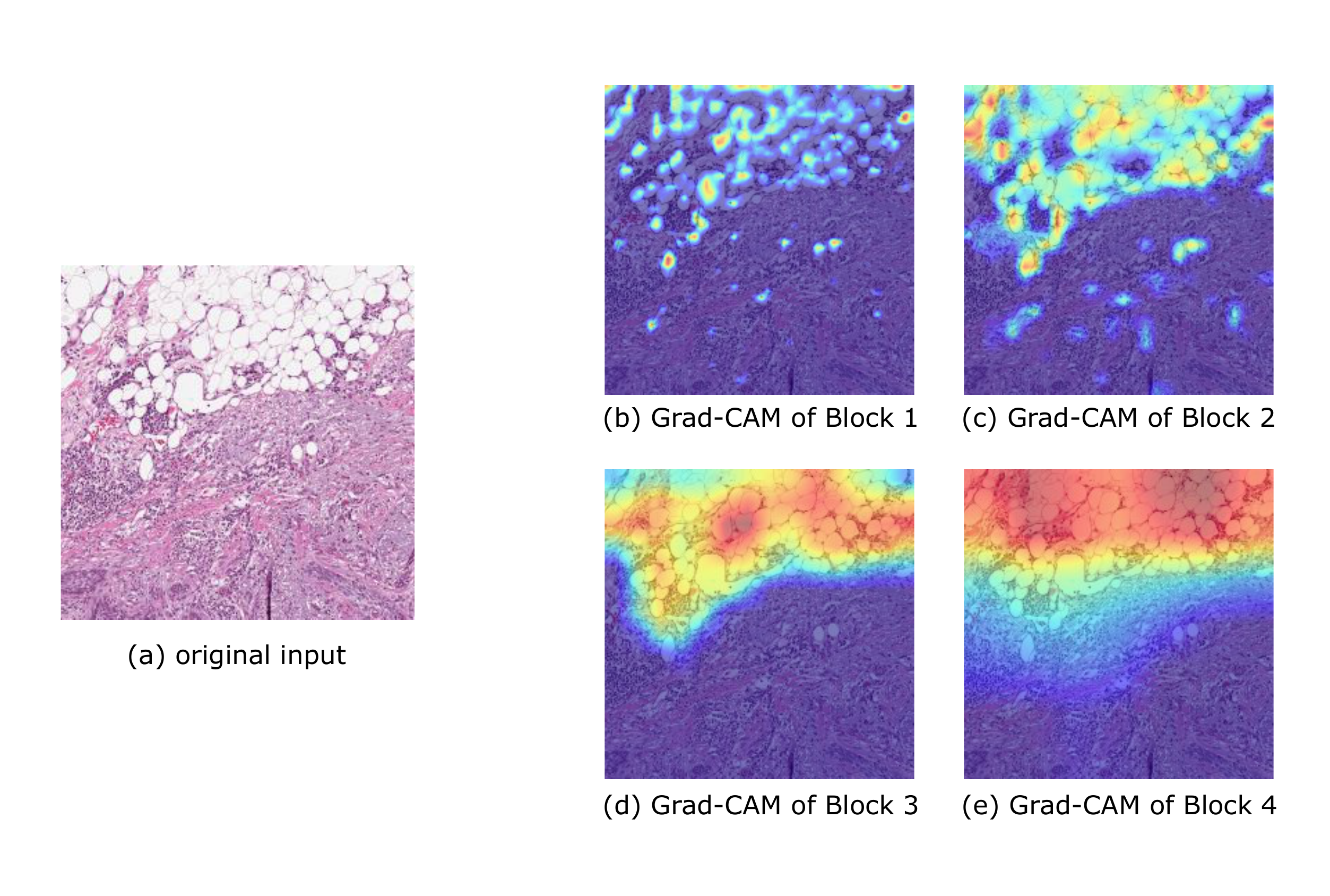}
\caption{Grad-CAM visualisations of the four residual blocks in ResNet18 trained to detect tumours in skin WSIs. In the example, the model's prediction of the absence of a tumour is incorrect: tumour cells are present in the bottom of the patch (image (a)). Observing Grad-CAM visualisations (images (b) - (e)), where high prediction importance is shown in red, we can conclude that throughout the blocks the NN is incorrectly using the presence of fat tissue to indicate absence of tumour; an indication of potential bias in the trained model.}
\label{fig:4BlocksGradCam}
\end{figure}

Visualisations on an input image methods, such as Grad-CAM, can be used to increase the transparency as well as uncover some potential biases that the model has. Figure \ref{fig:4BlocksGradCam} shows an example of such a case. We trained a ResNet18 neural network to predict if a patch from a WSI of skin contains a tumour. ResNet18 contains four residual blocks, units consisting of several layers. It is connected with the previous block with a skip connection \cite{HeResNet}. We used the Grad-CAM technique to visualise each of the four blocks. The resulting heatmaps show which pixels that are most important for the prediction. This provides an overview of how the attention of an NN is changing through the layer blocks when the image is classified. The model predicted that there is no tumour, but the patch does contain tumour cells in the lower part of the image. The Grad-CAM method uncovers a potential bias: the model uses fat tissue as an indication of the absence of the tumour cells, which could be caused by an over-representation of fat tissue in patches without a tumour in the training data set.
\par
All methods described so far aim to provide a deeper understanding of an NN trained to do classification. However, AI is not only used for classification tasks. Wu and Song \cite{Wu_2019_ICCV} has proposed how to improve the interpretability of a CNN trained for object detection. They create an architecture of an NN based on R-CNN that provides not only the classification score but also the bounding box on the region of interest (the target object). 
Furthermore, XAI visualisation on the input image can increase the interpretability of the segmentation task. Kwon et al.  \cite{Kwon2020} have used a technique for estimating uncertainty first described by Kendall and Gal \cite{KendallGal2017} and created heatmaps of uncertain regions of the segmentation. A large area of uncertain regions can warn a user that the NN is making a poorly informed guess.

\subsubsection{Showing by example}
The methods in this group are diverse but share the foundation that their explanations are based on presenting and discussing examples, either from the original data or from auxiliary data. This is a small group of methods so far, but it has great potential as some research claims that this type of explainability may be the most intuitive for a human user \cite{Mittelstadt2019}.
\par
The first subgroup of methods uses examples from the original data set in order to understand an NN better. Lapuschkin et al. \cite{Lapuschkin2019} provide a visualisation of clustering input images that an NN determined to be from the same class. The clustered images reveal potential rationale for how an NN assigns samples to a certain class. For example, such exploration may reveal that an image is classified as containing a horse if there is a white fence, if there is a person in a certain position or if there is a clearly visible horse head. Similar results are achieved in work by Kevin et al. \cite{Kevin2018} where authors grouped WSIs of neuropathological tissue that an NN perceived to be similar. Moreover, at each layer of an NN, we can compare the representations of the input image to the representations of all other images in the data set and see what labels have the $k$ most similar other images \cite{Papernot2018}. This may help to detect out-of-distribution images. Alsallakh et al. \cite{Alsallakh2018} explore a confusion matrix produced by an algorithm in order to determine which classes that are causing most trouble and the possible reason for the confusion.  
\par
\begin{figure}[t]
\centering
\includegraphics[width=\textwidth]{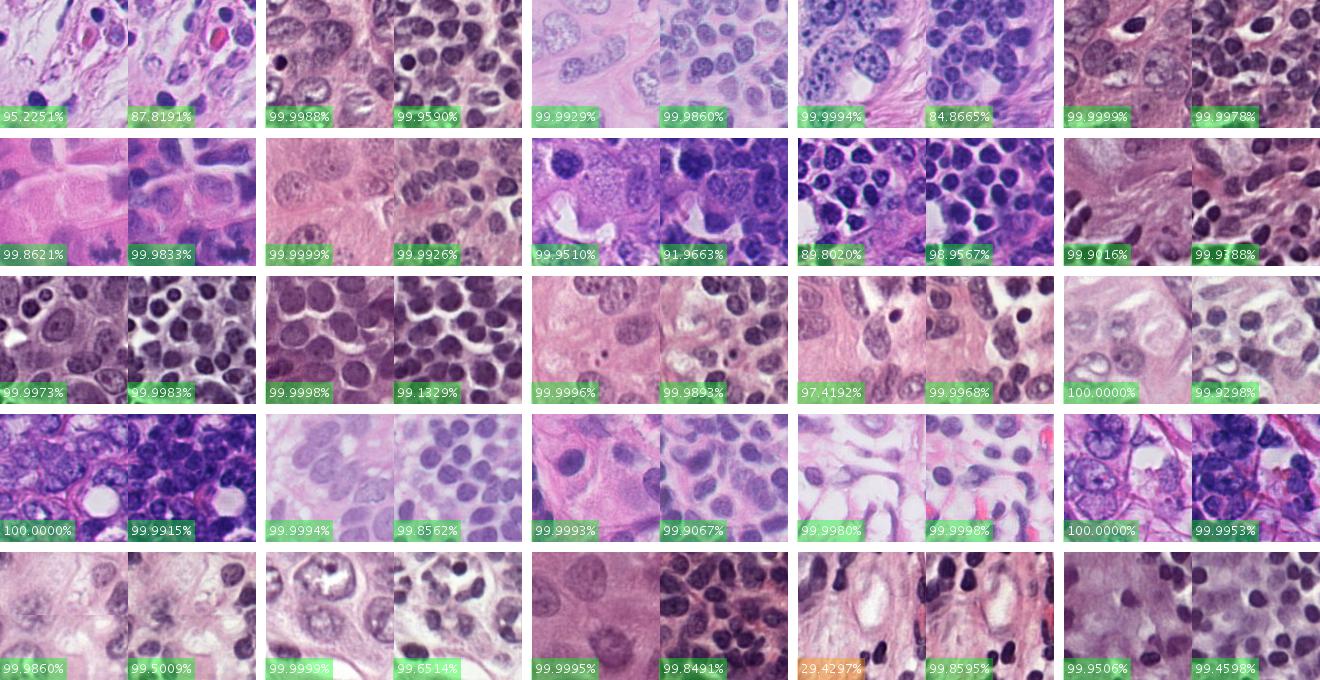}
\caption{Cycle-GANs transform a WSI patch with tumour to a synthetic healthy patch. Such counterfactual examples could boost pathologists' confidence that a tumour prediction is correct. The illustration consists of 25 pairs of patches: to the left are original patches with tumour and to the right are their respective transformation that illustrates how the patch could look if it were healthy. The percentage shows how confident the target NN is that the left patch contains tumour and the right patch is healthy, respectively.}
\label{fig:C_GANs}
\end{figure}

Other methods show examples that do not come from the original data set. Seah et al. \cite{Seah2018} proposed an idea that if we generate the most similar image to the original one but that would be classified to a different class by an NN, we could compare these images and understand which parts or differences are used by the NN to predict the correct class. Figure \ref{fig:C_GANs} illustrates such a method used on a tumour classifier. We trained a Cycle-GAN \cite{Zhu_2017_ICCV} to transform patches that contain tumour cells to healthy ones. The score in the left corner of each image shows the prediction score for each patch by the NN under scrutiny; the high confidence of the NN proves that the transformation was successful. Such counterfactual illustrations could capture and convey complex explanations of diagnostic predictions.
\par
Furthermore, some researchers explore if human-comprehensible concepts, such as striped patterns, gender, or a tie, are influencing an NN \cite{Bau2017,Kim2018}. These methods require a separate data set that contains the images with the concepts of interest. They can show which images of a certain concept that are most relevant to a chosen class that an NN is predicting. Also, they can reveal which particular images in the original data set that are influenced by the concept most. For example, they can show images with the striped patterns (from the concepts data set) that are most related to a Chief Enterprise Officer (CEO) class in an NN and which images labelled CEO (from the original data set) are mostly influenced by the stripe patterns. This knowledge could potentially alleviate biased predictions in an AI solution.

\subsubsection{Auxiliary measures}
This subsection provides an overview of various measures that have been developed for understanding an NN better. They do not necessarily have anything in common apart from the main aim to make an NN more transparent. However, all of them explore an important way of representing the results: providing an informative score or measure.
\par
Uncertainty measures give insight into how much we can trust the outcome of an AI solution. These measures can be used i) to construct prediction intervals that show how much the prediction could vary \cite{Pearce2018PredictionIntervals},  ii) to create heatmaps as discussed in Section \ref{Heatmaps} above, iii) to create plots illustrating uncertainty \cite{Ayhan2018}, or iv) to be presented as a score to the user \cite{Gal2016}.
\par
Other scores provide a measure of importance that helps to understand how an NN makes the prediction. For example, Koh and Liang \cite{WeiKoh2017} assigns an importance score for each training image by exploring how the predictions of the target NN would change if the particular image would be removed from the training data. 

\subsection{Technical approach}
This subsection gives an insight into how the explainability methods work technically. Thus, the categorisation is probably more relevant for readers with a technical background. Knowing what strategies that have been commonly applied in acquiring the visualisations or the scores may provide a good starting point for identifying what technical approaches that have not yet been explored. It is important to note that we do not aim to describe the technical details of each method thoroughly, instead, we provide a general overview of what kind of techniques researchers have used so far.

\subsubsection{Activation optimisation}
All methods in this subsection use an optimisation algorithm on the activations of the target NN in order to create their visualisations. There are two distinct ways of how optimisation in the context of neural network understandability may be used. Some techniques aim to explore the inner workings of an NN by finding the patterns that maximise the activation of a chosen neuron or a combination of neurons. The main idea can be illustrated by the following optimisation problem:

\begin{equation}
        x^{*} = \argmaxA_{x} h_{i,j}(\theta, x) + regulariser,
\end{equation}
where $h_{i,j}$ is the activation of the target neuron with indices $i,j$, that has been computed from the input sample $x$.

$\theta$ is the set of parameters used for computing the activation and $x^{*}$ is the estimated sample that maximises the output the neuron at $i,j$. This objective is usually achieved using gradient descent, and most proposed activation maximisation methods mainly differ in what regulariser they propose to use \cite{Carter2019, Erhan2009, Nguyen2016, Olah2018, yosinski2015understanding}.
\par
Optimisation techniques are also used for inverting NNs. Mahendran and Vedaldi \cite{Mahendran2016} proposed a method to reconstruct the original image from the activations of a chosen layer of an NN. Given that a neural network is some composite function  $\Phi_{0} = \Phi(x_{0})$ with input image $x_0$, they acknowledge that finding an inverse of $\Phi_{0}$ is difficult; neural networks are composed of many non-invertable functions. Hence, this method aims to find an image whose representation best matches the target input image by reducing the loss between $\Phi(x)$ of some input $x$ and the target $\Phi_{0}$. There are a few other efforts where model inversion tools also are proposed to achieve a better understanding of an NN \cite{Dosovitskiy2016,Mahendran2015}.

\subsubsection{Back-projection to input space}
Back-projection methods reflect a prediction score of an NN or activation of a target neuron back to the input space. They generate the patterns that triggered the neuron or create sensitivity maps over the input pixels. We describe this category by highlighting two representative approaches, deconvolution and saliency map. 
\par
Deconvolution is a technique that inverts the direction of the information flow in an NN \cite{ZeilerFergus2014}. In a higher layer a target neuron is chosen and the activations of all the other neurons in that layer are set to 0. Then, the activation of the target neuron is passed back through the layers to the input space. In order to invert the max pooling layers which are normally non-invertable, Zeiler and Fergus \cite{ZeilerFergus2014} propose to compute so-called switches during the forward pass: they record which positions that contained the maximum value within each pooling region. The resulting signal in the input layer is used to construct an image which shows what patterns that triggered the activation of the target neuron. Guided Backprop \cite{Springenberg2014} is another method also built on this approach.
\par
Saliency map is a technique that uses Taylor expansion in order to back-project a prediction score to the input space. The technique is based on the idea that the prediction of a neural network can be approximated by a linear function:
\begin{equation}
       S_{c}(x) = w^{T}x + b, 
\end{equation}
\noindent where $S_{c}$ is the class score, $w$ and $b$ are the weight vector and bias of the model, respectively, and $x$ is the input image. Then the first-order Taylor expansion is used to determine the approximation for a weight vector $w^{T}$. The approximated $w^{T}$ reflects how much each pixel in the input contributed to the final score \cite{Simonyan2013}. There are a few other XAI methods that achieve back-projection to input space by using Taylor expansions \cite{Bach2015, Montavon2017} or other linear approximations that distribute the target score among the input pixels \cite{Selvaraju2017, Zhou2016}.

\subsubsection{Input perturbations}
Explainability can be achieved by inducing variations of the input. 
Methods in this category use perturbations on the input images in order to determine the important pixels for a prediction, or to estimate the uncertainty. It is important to note that the methods under this category may or may not analyse each perturbation separately. An example of such an algorithm is Local Interpretable Model-agnostic Explanations (LIME) \cite{Ribeiro2016}. The target input image is split into $k$ superpixels. A Ridge regression is then trained on $k$ perturbed images (each time only one superpixel is non-black) to predict the black-box classifier’s prediction score of that particular input. The parameters of the Ridge regression are used to determine which superpixel is the most important for the correct prediction. 
\par
Another method is using Shapley values for the explanation of a classifier \cite{LundbergLee2017}. Normally, these are computed by rerunning the classifier as many times as we have features (pixels in our case) and excluding one feature at a time. This provides an insight to which features that are important for the classification. However, such a procedure becomes very expensive with deep learning, hence the authors have offered several approximations for computing Shapley values for image classification. Similar use of perturbations or approximations of perturbations can be found in other XAI methods \cite{Ayhan2018,Dabkowski2017, Fong2017, Zintgraf2017}.

\subsubsection{Interpretable network design}
Another approach to explainability is to modify the architecture of an NN in order to arrive at models that produce more interpretable feature maps. This means that visualising them may give an insight into what objects a unit of an NN is detecting when making a classification decision.
\par
Interpretable CNNs \cite{ZhangWuZhu2018} is a method modifying the CNN filters; a special loss function is introduced that enforces the filters to capture human-interpretable objects better. The idea is that  visualising the receptive field of such filters could reveal behaviour of NN units, such as using cat head depictions to label the image as a cat. Another example is Interpretable R-CNN \cite{Wu_2019_ICCV}. Here, an architecture of an NN based on Faster R-CNN is created that provides not only the classification score but also the bounding box on the region of interest (the target object). This is achieved by introducing a new layer in the model called terminal-node sensitive feature maps that is based on graph theory. 

\subsubsection{Frequentist techniques} \label{FreqTech}
The methods in the group of frequentist techniques are built for standard neural networks. Therefore, applying such techniques usually does not require heavy modifications to the algorithm or the training pipeline. The methods are based on frequentist statistical theory and there are three main techniques commonly used: ensembles, bootstrapping and quantile regression.
\par
Ensemble methods use multiple learning algorithms to obtain a better predictive performance compared to the learning algorithms alone. Recently it has been proposed to generate the prediction intervals via ensembles of NNs \cite{Lakshminarayanan2017,Pearce2018PredictionIntervals}, i.e. utilising the variation between different models as a measure of uncertainty.
\par
Bootstrapping is a technique that uses random sampling with replacement in order to approximate the population distribution with the sample distribution. This provides means to estimate statistical measures such as standard error, bias, variance, and confidence intervals \cite{EfronBootstrap}. Osband et al. \cite{Osband2016} developed an efficient and scalable method that enables generation of bootstrap samples from a deep neural network to estimate the uncertainty.
\par
Finally, quantiles in statistics are dividing the range of the probability distribution or the sample distribution into continuous intervals that have equal probabilities \cite{YoungStatisticalInference}. Quantile regression, methods for estimating quantiles, can be used to build confidence intervals around the target variable (for example, the prediction of the NN) which enables estimation of the uncertainty. Tagasovska and Lopez-Paz \cite{Tagasovska2018} have proposed to implement an additional output layer called Simultaneous Quantile Regression for any deep learning algorithm. Simultaneous Quantile Regression would be trained on a special loss function that learns the conditional quantiles of a target variable. 
\par
It is also worth mentioning that while some of the methods described in this section are developed for regression problems, it is also possible to adapt them for classification tasks.

\subsubsection{Bayesian neural networks} \label{BNNs}
Bayesian Neural Networks (BNNs) are designed to incorporate uncertainty about the networks' parameters, weights and biases. Unfortunately, BNNs are computationally more complex than conventional NNs, require a different training pipeline, and may not scale well to big data sets nor deep architectures that are required for state-of-the-art performance \cite{Pearce2018BayesianEnsembling}. While it is out of scope for this survey to rigorously describe the research on BNNs, we next provide a brief overview of some key approaches. 
\par
Many researchers have worked on producing approximations of the intractable posterior distributions of BNNs -- approximations that can be used in practise. This includes, among others, applying Markov chain Monte Carlo methods \cite{Neal1995} and variational Bayesian methods \cite{Blundell2015}. Another challenge in training BNNs is performing the backpropagation, as there is a large number of possible values of the parameters. A few of the efforts have focused on improving the backpropagation algorithm for BNNs by introducing alternative algorithms, namely probabilistic backpropagation \cite{Hernandez-Lobato2015}, Bayes by Backprop \cite{Blundell2015}, and natural-gradient algorithms \cite{Khan2018}. Essential work has been done on determining the best way of finding a good prior as it strongly influences the quality of the uncertainty measurement \cite{Hafner2019}. Finally, some work focused on specifically developing Bayesian convolutional neural networks \cite{gal2015bayesian, shridhar2019comprehensive}, while others developed ways to distribute the estimated uncertainty between epistemic and aleatoric parts in BNNs \cite{Depeweg2018, KendallGal2017, Kwon2020}.

\subsubsection{Bayesian approximations} \label{BayesApprox}
Bayesian theory can be applied also to standard NNs to estimate uncertainty. In this way, uncertainty awareness can be introduced without having to tackle the shortcomings of BNNs. They use derivations and modifications to a standard NN training in order to incorporate the Bayesian inference.
\par
Pearce et al. \cite{Pearce2018BayesianEnsembling} showed how to add a Bayesian prior via regularization of the parameters and train an ensemble of standard NNs in order to approximate the posterior distribution. Postels et al. \cite{Postels2019} proposed to reduce the computational complexity of ensembles and other resampling-based methods by adding noise to the parameters of an NN during training. Some researchers focused on how to derive Bayesian inference approximations using some common techniques in modern NN training, such as dropout \cite{Gal2016} and batch normalisation \cite{Teye2018}. Finally, Ritter et al. \cite{Ritter2018} applied a Laplace approximation to obtain uncertainty estimates after an NN is trained.  

\subsubsection{Other techniques}
A number of methods base the generation of explanations on techniques that do not fall into any of the previous categories. Some methods use Generative Adversarial Networks (GANs) as part of their XAI solution \cite{NguyenDosovitskiy2016,Seah2018} (similar to Figure \ref{fig:C_GANs}). Moreover, uncertainty estimates can be based on conformal prediction; a technique that determines the confidence of a new prediction based on the past experience \cite{Papadopoulos2007, Papernot2018}. The training data set can be explored using influence functions, a classic technique from statistics, which help to detect the points most contributing to a given prediction \cite{WeiKoh2017}. Linear binary classifiers and binary segmentation algorithms have also been used to determine to what concepts (defined by a separate data set) a target NN is responding \cite{Bau2017, Kim2018}. Finally, techniques for dimensionality reduction, such as t-distributed Stochastic Neighbour Embedding (t-SNE)~\cite{Maaten2008}, makes it possible to reduce the dimensionality of feature representations of an NN and highlight which input images the model is perceiving to be similar \cite{Kevin2018}. It is also possible to extract information on how different NNs relate to each other by considering the weights of a large number of trained models~\cite{Eilertsen2020}.

\subsection{XAI methods in medical imaging}

In this final subsection of the XAI method overview, we highlight some previous efforts of XAI techniques specifically addressing the medical imaging domain. 
\par
Palatnik de Sousa et al. \cite{PalatnikdeSousa2019} explored an AI algorithm that classifies lymph node metastases by creating heatmaps of the area in the input patch that contributed most to the prediction. The authors found that deep learning algorithms trained for this task have underlying ‘reasoning’ behind their predictions similar to human logic. Similar findings are reported in a study on classifying Alzheimer’s disease \cite{TangChuang2019}. Huang and Chung \cite{Huang2019} addressed the need for an informed decision by training a CNN model to predict the presence of cancer for a given WSI. At the test time of this algorithm, XAI methods that explain the predictions by creating visualisations on the input are used to detect the areas of the tissue that are unhealthy. It provides ‘evidence’ why this WSI should be classified as unhealthy as well as localises the cancerous cells. The authors showed that the detected areas closely correlated with the pathologists labelling. Therefore, it provides a meaningful intuition of why AI categorised the whole slide as containing a tumour. 
\par
Incorporating uncertainty measures in AI solutions for digital pathology could help to increase transparency as well. Kwon et al.  \cite{Kwon2020} has shown that the highly uncertain regions in ischemic stroke lesion segmentation often correlate with the pixels that are incorrectly labelled by the algorithm. This means that uncertainty measures provides means for a doctor to spot possibly wrong predictions and understand when he or she should be cautious of the AI decision. Fraz et al. \cite{FrazShaban2018} fulfilled this idea by incorporating an uncertainty measure in their model, and showed that the quality of microvessel segmentation did indeed improve. 
\par
These results are inspiring and demonstrate the potential of XAI techniques in medical imaging. Nevertheless, the current body of work is still quite limited. There is a need for more in-depth research on applied XAI methods in the domain, and particularly so in digital pathology, as will be further discussed in the next section.

\section{Open Problems} \label{Overview}
There are several open problems arising with XAI application for digital pathology. The stakeholders involved in XAI development and usage should be aware of the potentially misleading explanations as well as the lack of ways to evaluate different XAI methods.
\par
The first problem is caused due to the design of the explanations and their user interactions. To increase human understanding, it is important to formulate causal explanations, that is, why the AI algorithm made that prediction. Holzinger et al. \cite{Holzinger2020} argue that analysing how an AI solution arrived to the prediction may not always result in a satisfactory explanation as a more thorough understanding of how humans interpret and use explanations is needed. This point is also highlighted by Mittelstadt et al. \cite{Mittelstadt2019}, further detailing that explanations may become damaging if they are phrased confusingly or do not match the users' expected format. 
\par
Another pitfall is if the explanation designer has a certain agenda. As there may be a strong incentive to promote trust in the AI predictions, explanation design runs the risk of being more persuasive than informative, as demonstrated in the work by Kaur et al. \cite{Kaur2020}. This could have especially severe consequences in XAI solutions developed for digital pathology, as overconfidence in them could result in patient hazards, as well as in a setback for the needed development of AI assistance to improve diagnostics.
\par
Even though XAI tools could increase the understandability of an AI solution, so far we lack a solid scientific evaluation framework that would allow us to understand when they work well and what limitations they have. This challenge arises due to the fact that usually the ‘ground truth’ is unknown for most of the outputs by any XAI technique. There are a few works attempting to answer these questions with some studies showing alarming results that the assessed methods do not always live up to the expectations \cite{ AdebayoGilmer2018, fang2019evaluating, Nie2018ATE,Samek2017, SnoekOvadia2019}. The question marks about the performance and evaluation of XAI methods remain for both the general case and specifically for digital pathology. 

It is often stated that AI algorithms are black boxes that need to become transparent. While this is an illustrative metaphor, it is also necessary to carefully consider what type of transparency that is meaningful and effective in different scenarios and for different stakeholders. In summary, there is strong rationale for XAI advances tailored for digital pathology, however the challenge of constructing meaningful explanation and evaluating the performance of a chosen XAI method still remains. 

\section{Future outlook} \label{Discussion}
Our analysis of the overview presented in the previous sections has led to several key findings that can be informative for future research in XAI for digital pathology. A first insight is that the comprehensive list of identified explainability scenarios points to the fact that there are many techniques to consider when developing XAI solutions in the domain. XAI is also an important concept that needs to be incorporated for decision support -- due to the potentially high costs of errors in healthcare, pathologists are concerned about using black-box algorithms in their daily practices and call for an increased transparency \cite{MolinWoundefinedniak2016}. Moreover, there is a substantial heterogeneity both in terms of the desired benefits that AI tools should bring, and the types of prediction tasks to be performed. Thus, it appears clear that even within this niche a multi-faceted XAI toolbox will be needed. 
\par
In order for the XAI researcher to navigate the digital pathology landscape, it is valuable to consider the three usage scenarios: development, QA, and diagnostic work. For example, for the model developer, all of the explanation types would be relevant, whereas the inner workings of the neural network ought to be of little relevance for QA or diagnostics. Understanding training data quality is probably unnecessary for diagnostic work but may be quite important for the QA specialist to assess limitations of the model at the specific lab.
\par
Similar mappings can be done with respect to result representation. A likely difference between usage scenarios here is that synthetic visualisations would only be valuable to the model developer. There may, however, be exceptions. Mittelstadt et al. \cite{Mittelstadt2019} argue that counterfactual and contrastive explanations are suitable for intuitive human understanding. The synthesised counterfactual image proposed by Seah et al. \cite{Seah2018} is an interesting direction, combining the synthesis and showing by example explanations. It is our expectation that combinations of XAI techniques will be needed in order to be sufficiently effective in most digital pathology applications.
\par
Finally, this survey sheds some light on the role of uncertainty in relation to XAI. Whereas uncertainty estimation sometimes is seen as a separate topic, the survey indicates that uncertainty is an integral part of the XAI scope when seen from an end-user perspective. Our review lists some useful existing methods for performing and evaluating uncertainty estimation. There is, however, ample room for further research efforts, not the least directed towards imaging diagnostics applications. Moreover, we argue that AI solutions for clinical routine need to have some level of uncertainty awareness. While this is not part of bread-and-butter AI development today, we hope that incorporating uncertainty soon will be the standard, which in turn will have a direct positive impact for broad XAI deployment.


\bibliographystyle{splncs04}
\bibliography{bibliography.bib}

\par
\pagebreak

\appendix
\section{Reviewed Methods}\label{sec:appendix}
\par
\begin{footnotesize}
\setlength{\LTleft}{-20cm plus -1fill}
\setlength{\LTright}{\LTleft}
\begin{longtable}{|l|l|l|l|}
\hline
 \textbf{Method} & \textbf{Explains} & \textbf{Result representation} & \textbf{Technical approach}  \\ \hline
 Quantile Regression & Irreducible & Auxiliary measures & Frequentist techniques \\
 Forests \cite{Meinshausen2006} & uncertainty & & \\ \hline
 Conformal Prediction \cite{Papadopoulos2007}  & Reducible uncertainty & Auxiliary measures & Other techniques \\ \hline
 Activation  & Inner workings & Synthetic visualisations & Activation optimisation   \\
 maximisation \cite{Erhan2009} & & & \\ \hline
 Saliency maps \cite{Simonyan2013} & Predictions & Visualisations on the input & Back-projection to   \\ 
  &  &  & input space  \\ \hline
 DeconvNet \cite{ZeilerFergus2014} & Predictions & Synthetic visualisations & Back-projection to   \\ 
  &   &  & input space  \\ \hline
 Guided Backprop \cite{Springenberg2014} & Predictions & Synthetic visualisations & Back-projection to   \\ 
  &  of NN &  & input space  \\ \hline
 Object detectors in & Prediction; &  Visualisations on the input & Input perturbations  \\ 
 Deep Scene \cite{Zhou2014} & Inner workings &  &   \\ \hline
 Layer-wise relevance  & Predictions  & Visualisations on the input  & Back-projection to   \\ 
 propagation \cite{Bach2015} &  &  & input space   \\ \hline
 Probabilistic &  Reducible uncertainty  & Auxiliary measures & Bayesian neural  \\ 
 backpropagation \cite{Hernandez-Lobato2015} &  &  & networks \\ \hline
 Inverting Deep Image & Inner workings & Synthetic visualisations & Activation optimisation  \\ 
 Representations  \cite{Mahendran2015} &  &  &  \\ \hline
 BCNN with Bernoulli & Reducible uncertainty & Auxiliary measures & Bayesian neural\\
  approximations \cite{gal2015bayesian} & & & networks \\ \hline
 Deep visualization \cite{yosinski2015understanding} & Inner workings & Synthetic visualisations & Activation optimisation \\ \hline
 Bayes by Backprop \cite{Blundell2015} & Reducible uncertainty & Auxiliary measures & Bayesian neural \\ 
 & & & networks \\ \hline
 Bootstrapped DQN \cite{Osband2016} &  Reducible uncertainty & Auxiliary measures & Frequentist techniques \\ \hline
 Multifaceted Feature & Inner workings & Synthetic visualisations & Activation optimisation  \\ 
 Visualization \cite{Nguyen2016} &  &  &  \\ \hline
 Inverting feature & Inner workings & Synthetic visualisations & Activation optimisation\\ 
 representations \cite{Dosovitskiy2016} &  &  & \\ \hline
 Class activation  & Prediction & Visualisations on the input & Back-projection to   \\ 
 mapping (CAM) \cite{Zhou2016} &  &  & input space  \\ \hline
 Dropout as Bayesian &  Reducible uncertainty & Auxiliary measures & Bayesian   \\
 approximation \cite{Gal2016} &  &  & approximations \\ \hline
 Disco nets \cite{Bouchacourt2016} & Irreducible uncertainty & Auxiliary measures & Bayesian approximations  \\ \hline
 Deep generator &  Inner workings & Synthetic visualisations & GANs methods \\ 
 networks \cite{NguyenDosovitskiy2016} &  &  &  \\ \hline
 Distinct class & Prediction & Visualisations on the input & Back-projection to \\ 
 saliency maps \cite{Shimoda2016} &  &  & input space \\ \hline
 LIME \cite{Ribeiro2016} & Prediction & Visualisations on the input & Input perturbations \\ \hline
 Uncertainty with & Reducible uncertainty; &  Auxiliary measures & Frequentist techniques   \\
 deep ensembles \cite{Lakshminarayanan2017} & Irreducible uncertainty  &  &   \\
 & of NN & & \\ \hline
 Deep Taylor & Predictions  & Visualisations on the input  & Back-projection to   \\ 
 decomposition \cite{Montavon2017} &  &  & input space   \\ \hline
 SHAP \cite{LundbergLee2017} & Predictions & Visualisations on the input & Input perturbations \\ \hline
 PatternNet and & Predictions & Visualisations on the input & Back-projection to \\ 
 PatternAttribution \cite{Kindermans2017} &  &  & input space \\ \hline
 Integrated Gradients \cite{Sundararajan2017} & Predictions & Visualisations on the input & Back-projection to \\ 
 & & & input space \\ \hline
 Network Dissection \cite{Bau2017} & Inner workings & Showing by example & Other techniques \\ 
 & & & \\ \hline
 Grad-CAM \cite{Selvaraju2017} & Predictions & Visualisations on the input & Back-projection to \\ 
 & & & input space \\ \hline
 Uncertainties in Bayesian & Reducible uncertainty; & Auxiliary measures  & Bayesian neural  \\ 
 deep learning \cite{KendallGal2017} & Irreducible uncertainty &  & networks \\ \hline
 Meaningful  & Predictions & Visualisations on the input  & Input perturbations \\ 
 Perturbation \cite{Fong2017} & & & \\ \hline
 SmoothGrad \cite{Smilkov2017} & Predictions & Visualisations on the input & Back-projection to \\ 
 &  &  & input space \\ \hline
 Real Time Image & Predictions & Visualisations on the input & Input perturbations  \\ 
 Saliency \cite{Dabkowski2017} &  &  &   \\ \hline
 Prediction Difference &  &  &  \\ 
 Analysis \cite{Zintgraf2017} & Predictions & Visualisations on the input & Input perturbations  \\ \hline
 Influence Functions \cite{WeiKoh2017} & Predictions & Auxiliary measures & Other techniques \\ \hline
 Interpretable CNNs \cite{ZhangWuZhu2018} & Predictions & Visualisations on the input & Interpretable design  \\ \hline
 Generative Visual  &  &  &  \\ 
 Rationales \cite{Seah2018} & Predictions & Synthetic visualisations & GANs methods \\ \hline
 Weight-perturbation & Reducible uncertainty & Auxiliary measures  & Bayesian neural  \\ 
 in ADAM \cite{Khan2018} &  &  & networks \\ \hline
 Uncertainty in Bayesian & Reducible uncertainty; & Auxiliary measures & Bayesian neural \\ 
 Deep Learning \cite{Depeweg2018} & Irreducible uncertainty &  & networks \\ \hline
 Monotone composite  & Irreducible uncertainty & Auxiliary measures & Frequentist techniques  \\
 quantile regression NN \cite{Cannon2018} &  &  &  \\ \hline
 Deep k-Nearest & Predictions; & Showing by example; & Other techniques  \\ 
 Neighbors \cite{Papernot2018} & Reducible uncertainty & Auxiliary measures  &   \\ \hline
 Laplace approximation for & Reducible uncertainty & Auxiliary measures  & Bayesian approximations  \\ 
 estimating uncertainty \cite{Ritter2018} &  &  &  \\ \hline
 TCAV \cite{Kim2018} & Inner workings & Showing by example & Other techniques \\ \hline
 Bayesian uncertainty in  & Reducible uncertainty & Auxiliary measures & Bayesian approximations \\ 
 batch normalized NN \cite{Teye2018}  &  &  &  \\ \hline
 Excitation Backprop \cite{Zhang2018ExcitationBackprop} & Predictions & Visualisations on the input & Back-projection to \\
 & &  & input space \\ \hline
 Prediction Intervals for & Irreducible uncertainty & Auxiliary measures & Frequentist techniques \\ 
 Deep Learning \cite{Pearce2018PredictionIntervals}&  &  &  \\ \hline
 Bayesian Ensembling  \cite{Pearce2018BayesianEnsembling} & Reducible uncertainty & Auxiliary measures & Bayesian approximations \\ \hline
 Blocks \cite{Olah2018} & Inner workings & Synthetic visualisations & Activation optimisation \\ \hline
 Class Hierarchy &  &  &  \\
 in CNNs \cite{Alsallakh2018} & Inner workings & Showing by example & Other techniques \\ \hline
 Data augmentation for   & Irreducible uncertainty & Auxiliary measures & Input perturbations  \\
 uncertainty estimation \cite{Ayhan2018} & & & \\ \hline
 Visualization and Anomaly  & Inner workings; & & \\
 detection using t-SNE \cite{Kevin2018} & Reducible uncertainty & Showing by example & Other techniques \\ 
 &&& \\ \hline
 Uncertainty via noise & Reducible uncertainty & Auxiliary measure & Bayesian neural \\ 
 contrastive priors \cite{Hafner2019} &  &  & networks \\ \hline
 FullGrad \cite{srinivas2019fullgradient} & Prediction & Visualisations on the input & Back-projection to   \\ 
  &  &  & input space  \\ \hline
 SpRAy \cite{Lapuschkin2019} & Inner workings & Showing by example & Other techniques \\ \hline
 Influence-directed & Predictions;  & Showing by example & Other techniques \\ 
 explanations \cite{Leino2019} & Inner workings  &  &  \\ \hline
 Bayesian CNN with & Reducible uncertainty & Auxiliary measures & Bayesian neural\\
 variational inference \cite{shridhar2019comprehensive} & & & networks \\ \hline
 Uncertainty quantification & Reducible uncertainty; & Auxiliary measures & Bayesian neural  \\ 
 in Bayesian NN \cite{Kwon2020} & Irreducible uncertainty &  & networks  \\ \hline
 Dissecting the weight & Inner workings & Auxiliary measures & Other techniques \\
 space of NN \cite{Eilertsen2020} & & & \\ \hline

\caption{List of methods and their categorisations, ordered by the year of publishing.} \label{tab:methods_cats}
\end{longtable}
\end{footnotesize}

\end{document}